\pdfoutput=1

\documentclass[11pt]{article}

\usepackage[]{emnlp2021}

\usepackage{times}
\usepackage{latexsym}

\usepackage[T1]{fontenc}

\usepackage[utf8]{inputenc}

\usepackage{microtype}

\usepackage{multirow}
\usepackage{graphicx}
\usepackage{subcaption}
\usepackage{color, colortbl}
\usepackage{adjustbox}
\usepackage{xltabular}
\definecolor{DarkGray}{gray}{0.8}
\usepackage{hyperref}

\title{Exploring Multitask Learning for Low-Resource Abstractive Summarization}

\author{Ahmed Magooda, Mohamed Elaraby, Diane Litman \\
        University of Pittsburgh \\ Pittsburgh, PA, USA \\ \texttt{\{aem132,mse30,dlitman\}@pitt.edu}}

\begin{document}
\maketitle

\begin{abstract}
This paper explores the effect of using multitask learning for abstractive summarization in the context of small training corpora. In particular, we incorporate four different tasks (extractive summarization, language modeling, concept detection, and paraphrase detection) both individually and in combination, with the goal of enhancing the target task of abstractive summarization via multitask learning. We show that for many task combinations, a model trained in a multitask setting  outperforms  a model trained only for abstractive summarization, with no additional summarization data introduced. Additionally, we do a comprehensive search and find that certain tasks (e.g. paraphrase detection) consistently benefit abstractive summarization, not only when combined with other tasks but also when using different architectures and training corpora.
\end{abstract}

\section{Introduction}
\label{sec:intro}
Recent work has shown that training text encoders using data from multiple tasks helps to produce an encoder that can be used in numerous downstream tasks with minimal fine-tuning (e.g., T5~\cite{raffel2019exploring} and BART \cite{lewis2019bart}). However, in multitask learning for text summarization, it is still unclear what range of tasks can best support  summarization, and most prior  work has incorporated only one additional task during training ~\cite{isonuma2017extractive,chen2019multi,pasunuru2017towards,gehrmann2018bottom}. Also, to our knowledge, no prior work has tried to tackle  multitask summarization in low-resource domains. 

\label{rquestions}Our work attempts to address these gaps by answering the following research questions: \textit{Q1) Can  abstractive summarization performance be boosted via multitask learning when training from a small dataset?} \textit{Q2) Are there some tasks that might be helpful and some that might be harmful for multitask abstractive summarization?} \textit{Q3) Will the same findings emerge if a very different learning model is used or if pretraining is performed?} \textit{Q4) Will the same findings emerge if a very different small training corpus is used?} To answer Q1, we use a pretrained BERT model \cite{devlin2018bert}  within a multitask framework, and train all tasks using a small-sized corpus of student reflections (around 400 samples). To answer Q2, we explore the utility of training on four different tasks (both alone and in combination) in addition to  abstractive summarization. To answer Q3, instead of fine-tuning with the BERT model, we perform experiments using the T5 transformer model \cite{raffel2019exploring}. To answer Q4, we replicate the student reflection  experiments using two very different corpora (news and reviews). Our results show that abstractive summarization in low resource domains can be improved via  multitask training. We also find that certain auxiliary tasks such as  paraphrase detection consistently improve abstractive summarization performance across different models and datasets, while other auxilary tasks like language modeling more often degrade model performance. 

\section{Related Work}

\textbf{Multitask learning.} Abstractive summarization has been  enhanced in multitask learning frameworks 
with one additional task, by integrating it with text entailment generation \cite{pasunuru2017towards}, extractive summarization
\cite{chen2019multi,hsu2018unified}, and  sentiment classification \cite{unified2020,ma2018hierarchical}. While other research has combined multiple tasks, 
\citet{lu2019multi}  integrated only predictive tasks, while  \citet{guo2018soft} used only generative tasks. Recently,  \citet{dou2021gsum}  proposed using different tasks as guiding signals. However, 
the guiding signals can only be used one signal at a time with no easy way to combine them. In contrast, our work focuses on both generative and predictive tasks, explores task utility in isolation and in all combinations, and demonstrates generalization of findings across multiple models and corpora. Furthermore, aside from the two auxiliary tasks \textit{(language modeling} \cite{magooda2020attend} and \textit{extractive summarization} \cite{pasunuru2017towards}) that have been examined before in the context of multitask summarization, we introduce two new additional auxiliary tasks \textit{(paraphrase detection, concept detection)(Section \ref{sec:aux_tasks})}. Finally, while previous work relied on large training corpora (e.g. CNN/DailyMail~\cite{hermann2015teaching}), we target low resource domains and try to overcome  data scarceness by using the same data to train multiple task modules. 

\textbf{Low resource training data.}
While most abstractive summarization work takes advantage of large  corpora such as CNN/DailyMail, New York Times, PubMed, etc. to train models from scratch \cite{hermann2015teaching,nallapati2016abstractive,cohan2018discourse}, recent work has also targeted low resource domains. Methods proposed to tackle little training data have included  data synthesis
 \cite{parida2019abstract,magooda2020abstractive}, few shot learning \cite{bravzinskas2020few}, and  pretraining \cite{yu2021adaptsum}. Our approach is different in that we use the same data multiple times in a multitask setting to boost performance.

\section{Summarization Datasets}

\textbf{CourseMirror (CM)}\footnote{https://petal-cs-pitt.github.io/data.html}\label{data:cmirror}
is a student reflection dataset previously used to study both extractive \cite{luo2015summarizing} and abstractive \cite{magooda2020abstractive} summarization. The dataset consists of documents (i.e., a set of student  responses to a reflective instructor prompt regarding a course lecture) and summaries from four course instantiations: CS, ENGR, S2015, and S2016.
\begin{table}[t]
\small
\begin{center}
\begin{tabular}{|l|c||c|c|c|}
\hline 
\textbf{Data} & \textbf{\# Docs} & \textbf{Train} & \textbf{Val} & \textbf{Test}\\
\hline
CS & 138 & 209 & 23 & 138 \\
\hline
ENGR & 52 & 286 & 32 & 52 \\
\hline
S2015 & 88 & 254 & 28 & 88 \\
\hline
S2016 & 92  & 250 & 28 & 92 \\
\hline
CNN-5\% & 2500 & 1500 & 500 & 500 \\
\hline
Amazon/Yelp & 160 & 58 & 42 & 60\\
\hline

\end{tabular}
\end{center}
\caption{\label{dataset_summary} Dataset summary.}
\end{table}

\textbf{CNN/DailyMail  (CNN-5\%)}\label{data:cnn}
 is a widely used summarization dataset consisting of around 300k news-oriented documents \cite{hermann2015teaching}. Since the focus of our research is low resource data, we randomly select 5\% (500 documents) from the CNN/DailyMail test and validation sets. Then, to keep the CNN-5\% data distribution similar to CM (3 courses for training, 1 for testing), we randomly sample 1500 documents for the training set.

\textbf{Amazon/Yelp\footnote{https://github.com/abrazinskas/FewSum}}
is a dataset of opinions~\cite{bravzinskas2020few} that is both small  as well as  similar to  CourseMirror in that documents consist of multiple human comments where order doesn't matter.
This dataset contains customer reviews from Amazon 
and Yelp 
of 160 products/businesses. For each of these, 8 reviews to be summarized are
 selected from the full set of reviews. 

Table \ref{dataset_summary} summarizes each dataset in terms of the number of documents 
and their distribution into training, validation, and test sets. The PDF appendix contains examples from  each dataset. 

\section{Proposed Models}
This section describes the different tasks used for multitask learning with the intuition behind them, followed by the two summarization models used. 
\subsection{Auxiliary Tasks}
\label{sec:aux_tasks}
\textbf{Extractive summarization (E)} aims to classify parts of a document (typically sentences) as either important (i.e. included in a summary) or not. It has  been used as  as an auxiliary abstractive summarization task~\cite{chen2019multi,hsu2018unified} as it can help the model focus on important sentences.

\textbf{Concept detection (C)}  detects important concepts (keywords) within an input text. Humans can have a general understanding of a topic's main idea by looking through concepts or keywords (e.g., keywords integrated into early pages of research papers or books). Thus, we hypothesize that this task can help the model focus more on major keywords.

\label{task:para}\textbf{Paraphrase detection (P)} aims to classify a pair of sentences as to whether they are conveying the same ideas using different wordings. We hypothesize that the relation between input documents and  summaries can be viewed as a potential paraphrasing. We use the MSRP paraphrase dataset \cite{dolan2005automatically}, in addition to summarization datasets, to  train a paraphrase detection task.

\textbf{Language modeling (L)}, in general, can help improve generative tasks. Training with LMs aims to skew the vocabulary slightly into the training data distribution.

\subsection{BERT Multitask Integration\footnote{https://github.com/amagooda/MultiAbs.git}}
\begin{figure}[htpb]
 \centering
    \includegraphics[width=.95\linewidth]{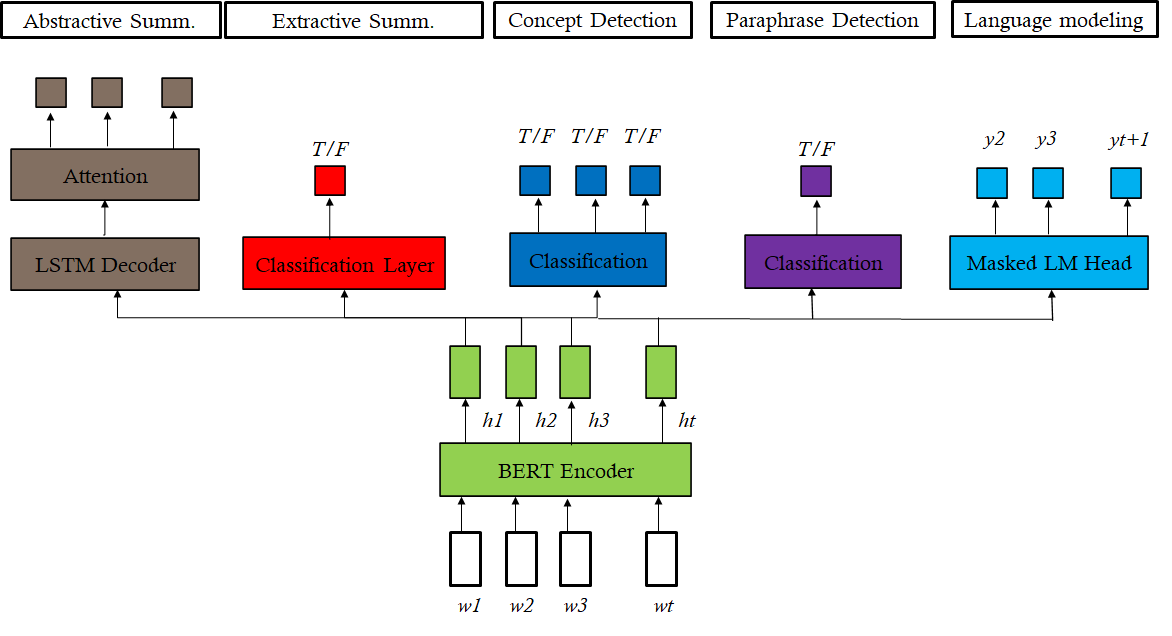}
    \caption{Proposed BERT-Multitask model.}
    \label{fig:model_fig}
\end{figure}

We use a pretrained BERT \cite{devlin2018bert} model as a shared sequence encoder followed by a set of different task-specific modules (Figure \ref{fig:model_fig}). In the {\bf single task} setting, only abstractive summarization is performed. In the {\bf multitask} setting (integrating one or more  auxiliary tasks), encoder weights are also fine-tuned alongside the rest of the model.

\textbf{Abstractive summarization (A)}. While  recent work often uses transformers to overcome issues of sequence length \cite{yan2020prophetnet},  LSTM based decoders consistently outperform transformer-based ones when trained from scratch on our small CM dataset. Thus, we use LSTMs for our abstractive summarization (primary) task.

\textbf{Extractive summarization (E)}: The model consists of a linear layer to classify a sentence as part of the summary or not. Document and input sentence are fed to BERT encoder in the format \textbf{[CLS] DW$_{1}$ DW$_{2}$...DW$_{n}$[SEP]SW$_{1}$ SW$_{2}$....SW$_{m}$}, where DW$_{i}$ is the i\textsuperscript{th} word of the input document, SW$_{i}$ is the i\textsuperscript{th} word of the sentence to classify, and ([CLS], [SEP]) are respectively the starting and separation tokens used by BERT.

\textbf{Concept detection (C)}: The module's objective is to classify each word within a sequence as either a part of a concept or not. The module consists of a fully connected layer following the BERT encoder. We prepare the data by extracting concepts using a  TF-IDF ranking algorithm \cite{thaker2019student}.

\label{task:para_2}\textbf{Paraphrase detection (P)}: The module consists of a fully connected layer classifier. Similar to extractive summarization, the input is passed to  BERT  in the format \textbf{[CLS]Sent$_1$[SEP]Sent$_2$}. Sent$_1$ and Sent$_2$ are the two input sentences for the MSRP dataset, and the input document and human summary for the summarization datasets.

\textbf{Language modeling (L)}: The language modeling module consists of a masked language modeling (MLM) attention head, fine-tuned using the MLM objective. Following the original BERT training from \citet{devlin2018bert}, input tokens are masked with probability 15\%, where masked tokens are either replaced by a special token (80\%), random word (10\%) or left unchanged (10\%).

\textbf{Model training}:
We train the model by training sub-modules consecutively. Thus, for each of the training epochs, we first train one of the sub-modules (e.g. abstractive) using the corresponding data batches, then we move to another sub-module (e.g. extractive), and so on. Each submodule is trained with Maximum likelihood estimation (MLE). We perform training using multiple optimizers. The intuition is to tune different modules with different rates. We tune the whole model using 3 optimizers: one for the BERT encoder, another  for the abstractive decoder, and the last  for the other modules. All optimizers are Adam optimizers, with different initial learning rates 5e\textsuperscript{-4}, 5e\textsuperscript{-3}, and 5e\textsuperscript{-5} for BERT encoder, abstractive decoder, and other modules respectively. We also performed experiments using a single optimizer for the whole model. Multiple optimizers consistently outperform a single optimizer.

\subsection{T5 Multitask Integration}
\label{sec:t5}
We also make use of the T5 \cite{raffel2019exploring}, which stores a large amount of knowledge about language and tasks. 
In the {\bf single task} setting, we fine-tune a pretrained T5 on the abstractive task ({\bf A}), using the low resource datasets.

\begin{figure}[htpb]
\centering
\begin{subfigure}{0.5\textwidth}
  \centering
  \includegraphics[width=.95\linewidth]{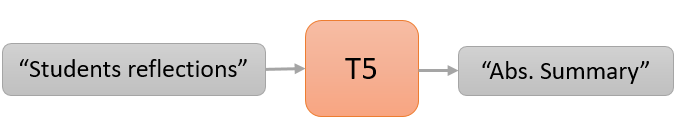}
  \caption{Fine-tune T5 on Abstracive Summarization Dataset}
  \label{fig:t5_abs}
\end{subfigure}\\
\begin{subfigure}{.5\textwidth}
  \centering
  \includegraphics[width=.95\linewidth]{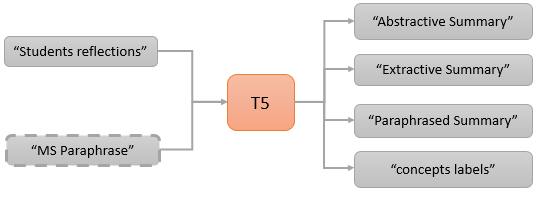}
  \caption{Fine-tune T5 on Mixture of tasks}
  \label{fig:t5_multi}
\end{subfigure}
\caption{Different fine-tuning conditions for T5. 
\textit{\textbf{(- -)} indicates optional additive data for Paraphrasing.}}
\label{fig:fig}
\end{figure}

In the {\bf multitask} settings, we adopt the T5 framework to train the mixture of tasks as text-to-text, which allows us to fine-tune in the same model simultaneously. Figure \ref{fig:fig} shows the settings used for training T5 model for both Single abstractive summarization task, and the multitask training with mixture of tasks. Since T5 is pretrained with CNN/DM, we don't perform experiments with CNN-5\% using T5. Also note that unlike BERT, T5 represents any task as language modeling. Thus, 
we dropped the language modeling auxiliary task for T5, as it would be a form of redundancy. 

\section{Experiments, Results and Discussion}
\begin{table}[!t]
\begin{center}
\small
\begin{tabular}{|l|c|c|c|}
\hline
Tasks & R1 & R2 & RL\\
\hline
Single task (A) & 26.82 & 4.71 & 21.5\\
\hline
A C & \cellcolor{DarkGray} 27.11 &  \cellcolor{DarkGray} 4.75 & 21.1\\
\hline
 A E   &  \cellcolor{DarkGray} 28.51  & \cellcolor{DarkGray} 4.91 & 21.41\\
\hline
{A P}  & \cellcolor{DarkGray} 27.83 & \cellcolor{DarkGray} 5.99 & \bf \cellcolor{DarkGray} 23.05\\
\hline
 A L   & \cellcolor{DarkGray} 27.22 & \cellcolor{DarkGray} 5.47 & 21.31\\
\hline
{A E L} & \cellcolor{DarkGray} 28.36 & \cellcolor{DarkGray} 5.62 & \cellcolor{DarkGray} {21.6}\\
\hline
{A E P} & \cellcolor{DarkGray} 27.68 & \cellcolor{DarkGray} 5.24 & \cellcolor{DarkGray}{21.81}\\
\hline
{A E C}  & \cellcolor{DarkGray}{27.41} & \cellcolor{DarkGray}{5.81} & \cellcolor{DarkGray}{22.13}\\
\hline
{A C P} & \bf \cellcolor{DarkGray}{29} & \bf \cellcolor{DarkGray}{6.43} & \cellcolor{DarkGray}{22.2}\\
\hline
 A L P & \cellcolor{DarkGray} 27.71 & \cellcolor{DarkGray} 5.82 & 21.14\\
\hline
 A L C & \cellcolor{DarkGray} 27.39 & \cellcolor{DarkGray} 6.09 & 21.36\\
\hline
 ALL & \cellcolor{DarkGray} 27.72 & \cellcolor{DarkGray} 5.55 & 21.31\\
\hline
\end{tabular}
\end{center}
\caption{\label{tab:BERTmodels_results} ROUGE results of {\it BERT} multitask on {\it CM}.  Gray indicates multitask R  is higher than single task score. \textbf{Boldface} indicates best R  across  tasks. {\it (Q1, Q2)}}
\end{table}

Our experiments evaluate performance using  ROUGE \cite{lin2004rouge} on F1. For CM data we  
report mean ROUGE using a leave-one-course-out validation\footnote{Individual course ROUGE scores are in  the Appendix.}, while for CNN-5\% and Amazon-Yelp we report ROUGE using held-out test sets.

\hyperref[rquestions]{\textbf{Q1}}: The gray cells in Table \ref{tab:BERTmodels_results} show that BERT multitask 
training for CM data can help improve single-task (A) training. For R1 and R2 we observe improvements across {\it all} task combinations.  While  some task combinations also improve RL ((A P), (A E L), (A E P), (A E C), (A C P)), others degrade performance, particularly when language modeling is involved (e.g., (A L), (A L P), (A L C), and (ALL)). Thus, while  multitask training can be effective, we  need to 
further explore task choice.

{\textbf{Q2}}: Prior work showed the utility of extractive summarization \cite{hsu2018unified} and language models \cite{magooda2020attend} as auxiliary summarization tasks, 
and we too observe similar behavior for R1 and R2 in Table~\ref{tab:BERTmodels_results}. For RL,  however, (A E) and (A L) failed to improve the score. 
Similarly, our new concept task (A C)  improves R1 and R2 but not RL.
On the other hand, integrating our proposed paraphrasing task (A P)  improves performance for all ROUGE scores. When we integrate two auxiliary tasks, (A E L), (A E P), (A E C), and (A C P)
improve all of R1, R2 and RL compared to single task performance.  For RL, it seems that adding E with another auxiliary task rather than in isolation  improves performance. Also, the (A C P) combination which uses  our two proposed tasks (concept, paraphrasing)  achieves the best R1, R2, RL in the 3-task setting.

\begin{table}[t]
\begin{center}
\small
\begin{tabular}{|l|c|c|c|}
\hline
Tasks & R1 & R2 & RL\\
 \hline
Single Task (A) & 36.08 & 10.94 & 31.57\\
\hline
A E & 29.99 & 8.80 & 24.80\\
\hline
A C & 35.46 & 10.76 & 30.81\\
\hline
{A P} & \bf \cellcolor{DarkGray}{36.75} & \bf \cellcolor{DarkGray}{12.13} & \bf \cellcolor{DarkGray}{32.30}\\
\hline
A C P & \cellcolor{DarkGray} 36.28 & \cellcolor{DarkGray} 11.59 & \cellcolor{DarkGray} 31.58\\
\hline
A E C & 29.19 & 8.69 & 25.20\\
\hline
ALL & 30.31 & 9.60 & 27.97\\
\hline
\end{tabular}
\end{center}
\caption{\label{tab:T5models_results} ROUGE results of {\it T5} (No language modeling auxiliary task) fine-tuned on {\it CM}. \textit{(Q3)}}
\end{table}

\hyperref[rquestions]{\textbf{Q3}}:  Table  \ref{tab:T5models_results} shows that some of the CM findings obtained using  BERT multitask  are similar when a  different model such as T5 is used for CM. Similar to BERT, incorporating paraphrasing into T5 helps improve all ROUGE scores when used as a single auxiliary task (A P) and in combination with the concept  task (A C P). On the other hand, the utility of (A E C) didn't transfer from BERT to T5.

\hyperref[rquestions]{\textbf{Q4}}: Shifting gears from changing the model to changing the data, Table \ref{tab:CNNBERTmodels_results} shows that when BERT multitask is applied to CNN-5\%,\footnote{Recall from Section~\ref{sec:t5} that T5 is not used for CNN-5\%.} there is now no task configuration that leads to improvement across all of R1, R2, and RL. However, the majority of combinations (6 of 11) improved two out of the three ROUGE scores, especially R2 and RL. Additionally, judging by ROUGE scores of certain combinations such as (A C) and (A P), we can see that the reduction in R1 (0.38, 0.47) is less than the improvements gained in R2 (0.39, 0.61) and far less than RL (2.05, 1.46) respectively. Thus, we can argue that paraphrasing auxiliary task tends to be very helpful either across different data or different models. 
To further verify the utility of paraphrasing across datasets, we also evaluated the T5 model\footnote{We only examined T5 since for CM, the T5 ROUGE scores  (Table~\ref{tab:T5models_results})
 were  higher than when using BERT (Table~\ref{tab:BERTmodels_results}).} 
on the Amazon/Yelp dataset. However, due to the lack of extractive annotation for Amazon/Yelp, we only examine  (A P), the best performing T5 combination for CM (Table~\ref{tab:T5models_results}). Table \ref{tab:T5models_results_2} shows that indeed paraphrasing is again helpful as an auxiliary task, as it improves all ROUGE scores for Amazon/Yelp.

\begin{table}[!t]
\begin{center}
\small
\begin{tabular}{|l|c|c|c|}
\hline 
 Tasks &  R1 &  R2 &  RL \\
\hline

Single Task (A) & 13.3  & 0.73 & 8.98\\

\hline
A C & 12.92  & \cellcolor{DarkGray} 1.12 & \cellcolor{DarkGray} 11.03\\

\hline
A E & 12.9  & 0.33 & 8.76\\

\hline
A P & 12.83  & \bf \cellcolor{DarkGray} 1.34 & \cellcolor{DarkGray} 10.44 \\

\hline
A L & \cellcolor{DarkGray} 13.43  & 0.65 & 8.36 \\

\hline
A E L &  \bf \cellcolor{DarkGray} 14.18 & 0.36 & \cellcolor{DarkGray} 10.1 \\

\hline
A E P & 12.82  & 0.64 & 8.53 \\

\hline
A E C & 11.52  & \cellcolor{DarkGray} 1.05 & \bf \cellcolor{DarkGray} 11.23 \\

\hline
A C P & 11.08  & \cellcolor{DarkGray} 1.09 & \cellcolor{DarkGray} 10.95 \\

\hline
A L P & 12.79  & 0.53 & 8.94 \\

\hline
A L C & 10.35  & 0.09 & \cellcolor{DarkGray} 9.81 \\

\hline
ALL & 11.15 & \cellcolor{DarkGray} 1.26 & \cellcolor{DarkGray} 10.49 \\
\hline

\end{tabular}
\end{center}
\caption{\label{tab:CNNBERTmodels_results} ROUGE results of {\it BERT} on {\it CNN-5\%}. \textit{(Q4)}}
\end{table}

\begin{table}[!t]
\small
\begin{center}
\begin{tabular}{|l|c|c|c|}
\hline
Tasks & R1 & R2 & RL \\
\hline
\citeauthor{bravzinskas2020few}& 36.25 & 9 & 22.36 \\
\hline
Single task (A) & 34 & 8.8 & 21.25 \\
\hline
    {A P} & \bf \cellcolor{DarkGray} 34.1 & \cellcolor{DarkGray}{\textbf{9.1}} & \bf \cellcolor{DarkGray} 21.7 \\
    \hline
\end{tabular}
\caption{\label{tab:T5models_results_2} ROUGE results of {\it T5} fine-tuned with paraphrasing on {\it Amazon/Yelp}. \textit{(Q4)}}
\end{center}
\end{table}

Finally, while the objective of our work is to explore the utility of auxiliary tasks across models and data, rather than to outperform the prior SOTA, we briefly compare our results to prior work where possible. For CM, multiple task combinations outperform the data synthesis method (CM + synthetic) from  \citet{magooda2020abstractive} on R2 and RL. For example, while (A C P) yielded 0.63 less R1, it had  0.98  and 1.52 higher R2 and RL, respectively.
For Amazon/Yelp, while our approach  increases R2 by 0.1 compared to \citet{bravzinskas2020few}, the R1 and RL scores are lower by 2.15 and .66, respectively.  These results show that there is  still room for improvement, particularly for R1, and suggest a future combination of our approach with such alternative low-resource methods.

\section{Conclusion and Future Work}
We explored the utility of multitask training for abstractive summarization, using three low resource  datasets (CM, CNN-5\%, Amazon/Yelp) 
and two fundamentally different models (BERT, T5) with different preconditions (i.e. BERT not pretrained with summarization dataset versus T5 pretrained with CNN dataset) to verify any observed behavior. We also integrated four different auxiliary tasks, in isolation and together. We conducted several experiments to find if training a multitask model, in general, is helpful, or if some tasks might introduce degradation in model performance. We showed that indeed some tasks might help improve ROUGE scores and some might not help, at least when trained in a low resource setting. 
We found that among all  task combinations, \textbf{(Abstractive + Paraphrase detection)} improved almost all ROUGE scores across  different datasets (CM , Amazon/Yelp, and CNN-5\%) and different 
models (BERT, and T5), 
with \textbf{(Abstractive + Concept detection + Paraphrase detection)} as another good candidate. We also found that  paraphrasing and concept detection, which had not been previously examined as auxiliary  abstractive summarization tasks, can be helpful for low resource data.
In the future, we plan to continue exploring the generality of our findings, by 
include new types of low resource data (e.g. discussions, emails), 
BART as one of the SOTA models, and new auxiliary tasks. We also plan to combine multitask learning with other low resource methods (e.g., data synthesis).

\section*{Acknowledgements}
The research reported here was supported, in whole or in
part, by the institute of Education Sciences, U.S. Department
of Education, through Grant R305A180477 to the University of Pittsburgh. The opinions expressed are those of the
authors and do not represent the views of the institute or the
U.S. Department of Education. We like to thank Khushboo Thaker for helping generating concepts, the Pitt PETAL group and the anonymous reviewers for advice in improving this paper.

\bibliography{MultiSumArxiv}
\bibliographystyle{acl_natbib}

\appendix

\section{BERT parameters}
In our BERT experiments we use the BERT basic uncased model which consists of 12 layers, and a hidden size of 768. We fine-tune the model using a single Nvidia P100 GPU for 85 epoch and a batch size of 4 and 8. The epoch with the highest ROUGE score on the validation set is used later for testing. We tried multiple initial learning rates, as different learning rates might be selected for different courses depending on the validation set performance. The multitask training is done in a sequential fashion, where during each epoch all tasks are trained sequentially (i.e. for each epoch, the abstractive sub-model is trained using all data, followed by the extractive sub-model, etc..). We use a maximum input length of (120, 200, and 250) tokens for CM experiments as the average document length of CM data is around 200 tokens, then used the most suitable length based on the validation set. We tried multiple max input lengths for CM as we noticed that there are repeated sentences within the reflections. So while smaller cut-offs like 120 can truncate some of the reflections (which can be repeated), it would lead to a faster training process. As for CNN-5\% we use a maximum of 500 (max is 512 for BERT). Shorter documents are padded and longer ones are truncated. We generate summaries using beam search with beams of length 5. The average length of CM summaries ranges from 35 to 42 tokens, and 56 for CNN. Thus we decided to limit the summary length to 50 tokens.

\section{T5 parameters}
 We use the \textit{3B T5} model, which is publicly available. The model consists of  $24$ layers for encoder and decoder. We set the initial learning rate to $0.001$, which the authors used in their summarization experiments. Due to the lack of hardware,  we couldn't perform \textit{Beam Search} decoding. We fine-tuned the course mirror data  on $7$ TPUs on Google Cloud for $5000$ steps.

\section{Data samples}
\subsection{CourseMirror (CM)}
Table \ref{tab:cmirror_summaries_example} shows an example of CM sample from CS course.
\begin{table*}[htpb]
\small
\begin{tabular}{|p{0.96\textwidth}|}
\hline \small\textbf{Prompt}\\
\hline \small Point of Interest (POI): Describe what you found most interesting in today's class.\\
\hline \small\textbf{Student Reflection Document}\\
\hline

\textbullet $ $ the dynamic bag\\
\textbullet $ $ I found the creation of the Bag to be the most interesting.\\
\textbullet $ $ Learning about bags was very interesting.\\
\textbullet $ $ Dr. Ramirez cleared up my understanding of how they should work.\\
\textbullet $ $ I was really interested in learning all about an entirely new data structure , the Bag.\\
\textbullet $ $ I 'm also noticing that as these classes get farther along , there is more focus on real world factors that determine strength of code like speed\\
\textbullet $ $ The bag concept was cool how basically acts like a bag in real life with its usefulness.\\
\textbullet $ $ Bags as a data type and how flexible they are.\\
\textbullet $ $ Discussing the Assignment 1\\
\textbullet $ $ I found the examples and drawings the teacher drew on the whiteboard the most interesting.\\
\textbullet $ $ Abstraction, though seemingly intimidating is kind of just giving programmers a break right?\\
\textbullet $ $ We 're given so many more abilities and operations without having to know exactly how to code that.\\
\textbullet $ $ That being said , while I understand the applications being explained to me , it 's hard to just manifest that on my own.\\
\textbullet $ $ Learning about resizing Bags dynamically\\
\textbullet $ $ The discussion of the underlying methods of ADTs such as bags was most interesting\\
\textbullet $ $ the implementation of an array bag\\
\textbullet $ $ Order does not matter when using a bag.\\
\textbullet $ $ It is important to keep all of the values in an array together.\\
\textbullet $ $ To do this , you should move an existing element into the vacant spot.\\
\textbullet $ $ Looking at ADT 's from both perspectives\\
\textbullet $ $ Information held in bags is not in any particular order\\
\textbullet $ $ different ways to implement the bag\\
\textbullet $ $ Thinking about a more general idea of coding with ADTs and starting to dig into data structures more specifically.\\
\textbullet $ $ Code examples of key concepts/methods is always helpful.\\
\textbullet $ $ I thought it was a good thing to go through the implementation of both the add ( ) and remove ( ) methods of the Bag ADT\\
\textbullet $ $ Today we were talking about a certain type of ADT called a bag.\\
\textbullet $ $ We talked about certain ways that we would implement the methods and certain special cases that we as programmers have to be aware of.\\
\textbullet $ $ If you were removing items from ADT bag , you can simply shift the bottom or last item and put it in the place where you we removed an item.\\
\textbullet $ $ This is because , in bags , order does not matter.\\
\textbullet $ $ Learning about managing arrays in a data structure\\
\textbullet $ $ The bag ADT and how it is implemented\\

\hline \textbf{Reference Abstractive Summary}\\\hline
 Students were interested in ADT Bag, and also its array implementation. Many recognized that it should be resizable, and that the underlying array organization should support that. Others saw that order does not matter in bags. Some thought methods that the bag provides were interesting.\\

\hline \textbf{Reference Extractive Summary}\\\hline
\textbullet $ $ Bags as a data type and how flexible they are.\\
\textbullet $ $ Thinking about a more general idea of coding with ADTs and starting to dig into data structures more specifically.\\
\textbullet $ $ I thought it was a good thing to go through the implementation of both the add() and remove() methods of the Bag ADT.\\ 
\textbullet $ $ Learning about managing arrays in a data structure.\\
\textbullet $ $ Information held in bags is not in any particular order.\\
\hline
\end{tabular}
\caption{\label{tab:cmirror_summaries_example} Sample data from the CourseMirror CS course.}
\end{table*} 

\subsection{Amazon/Yelp}
Table \ref{tab:amazon_summaries_example} shows an example of sample from amazon/Yelp data.
\begin{table*}[htpb]
\small
\begin{tabular}{|p{0.96\textwidth}|}
\hline
\bf Reviews\\
\hline
This pendant is so unique!! The design is beautiful and the bail is a ring instead of the typical bail which gives it a nice touch!! All the corners are smooth and my daughter loves it - looks great on her.I cannot say anything about the chain because used our own chain.:) Satisfied.\\

\hline  
It look perfect in a womens neck!! great gift, I thought for the price it was going to look cheap, but I was far wrong. It look great.Spect great reward from your woman when you give this to her; D\\

\hline
 The prettiest sterling silver piece I own now. I get so many compliments on this necklace. I bought it for myself from my hubby for Valentine's Day. Why not? When people ask where I got it, I simply say from my loving hubby. And he is off the hook as to what to get me. win + win.\\

\hline
I love hearts and I love 'love':) I do not have any negative feedback, the necklace is perfect and the charm is perfect. I just thought it would have been slightly bigger. Overall, I love my new heart necklace.\\

\hline
When I received the package, I was surprised and amazed because the necklace is so elegant, beautiful and the same as the picture shown here. I really love this necklace. It has a unique pendant designed. I will recommend it to someone to order it now...\\

\hline
Item is nice. Not a great quality item, but right for the price. Charm was larger than I expected (I expected small and elegant, but it was large and almost costume jewelry like). I think it is a good necklace, just not what I expected.\\

\hline
I got this as a present for my GF on Valintines day. She loves it and wears it every day! Its not cheap looking and it hasn't broken yet. The chain hasn't broken either even though it is very thin. Strongly recomend it!\\

\hline
Over all service has been great the only problem, I ordered a purple Mickey Mouse case for iPhone 4S they sent a black, n I felt it was to much trouble n such a small item to send back so needless to say its put back in a drawer somewhere\\
\hline
\hline
\bf Abstractive Summary\\
\hline
This silver chain and pendant are elegant and unique. The necklace is very well made, making it a great buy for the cost, and is of high enough quality to be worn every day. The necklace looks beautiful when worn bringing many compliments. Overall, it is highly recommended.\\
\hline
\end{tabular}
\caption{\label{tab:amazon_summaries_example} Sample data from the Amazon/Yelp data.}
\end{table*} 


\section{Full Results}
\label{sec:full_results_appendix}
Full results for BERT and T5 multitask models on CM data are shown in tables (\ref{tab:FullBERTmodels_results}, and \ref{tab:FullT5models_results}).
\begin{table*}[ht]
\begin{center}
\small
\begin{tabular}{|l|c|c|c|c|c||c|c|c|c|c||c|}
\hline
\multirow{2}{*}{ Tasks}   &  R1 &  R2 &  RL &  AVG & $\Delta$ &  R1 &  R2 &  RL  & AVG & $\Delta$ &Row\\
\cline{2-12} 
& \multicolumn{5}{|c|}{\bf CS0445} & \multicolumn{5}{|c|}{\bf ENGR}&\\
\hline
Single Task (A) & 26.93 & 3.98 & 21.04 & 17.32 & * & 27.19 & 7.27 & 22.66 & 19.04 & * & 1\\
\hline
A C & 27.09 & 4.85 & 20.12 & 17.35 & + & 30.14 & 7.67 & 22.96 & 20.26 & + & 2\\
\hline
A E  & 25.62 & 5.04 & 19.9 & 16.85 & - & 31.75 & 4.69 & 22.77 & 19.74 & + & 3\\
\hline
{\textbf{{A P}}}  & {\textbf{{28.13}}} & {\textbf{{7.13}}} & {\textbf{{23.45}}} & {\textbf{{19.57}}} & {\textbf{{+}}} & {\textbf{{28.56}}} & {\textbf{{7.29}}} & {\textbf{{23.99}}} & {\textbf{{19.95}}} & {\textbf{{+}}} & {\textbf{{4}}}\\
\hline
A L  & 25.53 & 4.69 & 21.48 & 17.23 & - & 30.04 & 7.36 & 24.27 & 20.56 & + & 5\\
\hline
A E L  & 28.18 & 6.48 & 21.34 & 18.67 & + & 33.75 & 8.64 & 26.86 & 23.08 & + & 6\\
\hline
A E P & 28.18 & 2.68 & 20.21 & 17.02 & - & 27.4 & 8.72 & 25.33 & 20.48 & + & 7\\
\hline
\bf {A E C}  & \bf {27.4} & \bf {6.58} & \bf {21.36} & \bf {18.45} & \bf {+} & \bf {28.87} & \bf {8.95} & \bf {24.33} & \bf {20.72} & \bf {+} & \bf {8}\\
\hline
\bf {A C P}  & \bf {28.18} & \bf {5.21} & \bf {20.67} & \bf {18.02} & \bf {+} & \bf {30.37} & \bf {10.84} & \bf {26.78} & \bf {22.66} & \bf {+} & \bf {9}\\
\hline
A L P  & 25.99 & 4.87 & 20.15 & 17 & - & 28.57 & 10.15 & 21.74 & 20.15 & + & 10\\
\hline
A L C & 32.15 & 5.42 & 21.99 & 19.85 & + & 25.81 & 7.66 & 21.51 & 18.33 & - & 11\\
\hline
\bf  ALL & \bf 28.34 & \bf 3.89 & \bf 22.79 & \bf 18.34 &\bf  + & \bf 28.54 & \bf 6.64 &\bf  25.7 & \bf 20.29 & \bf + & \bf 12\\
\hline
\hline
& \multicolumn{5}{|c||}{\bf S2015} & \multicolumn{5}{|c||}{\bf S2016}&\\
\hline
\bf Single Task (A) & 27.71 & 4.83 & 19.4 & 17.31 & * & 25.46 & 2.76 & 22.93 & 17.05 & * & 13\\
\hline
A C & 21.92 & 3.11 & 17.75 & 14.26 & - & 29.32 & 3.4 & 23.6 & 18.77 & + & 14\\
\hline
A E  & 27.99 & 5.07 & 20.97 & 18.01 & + & 28.7 & 4.87 & 22 & 18.52 & + & 15\\
\hline
{\textbf{{A P}}}  &  {\textbf{{28.6}}} & {\textbf{{4.84}}} & {\textbf{{22.33}}} & {\textbf{{18.59}}} & {\textbf{{+}}} & {\textbf{{26.03}}} & {\textbf{{4.7}}} & {\textbf{{22.43}}} & {\textbf{{17.72}}} & {\textbf{{+}}} & {\textbf{{16}}}\\
\hline
A L  & 26.12 & 4.43 & 18.37 & 16.31 & - & 27.22 & 5.4 & 21.14 & 17.92 & + & 17\\
\hline
A E L  & 23.44 & 4.35 & 18.72 & 15.5 & - & 28.09 & 3.01 & 19.51 & 16.87 & - & 18\\
\hline
A E P & 26.91 & 4.85 & 21.47 & 17.74 & + & 28.26 & 4.72 & 20.25 & 17.74 & + & 19\\
\hline
\bf {A E C}  & \bf {26.43} & \bf {4.45} & \bf {21.62} & \bf {17.5} & \bf {+} & \bf {26.94} & \bf {3.27} & \bf {21.24} & \bf {17.15} & \bf {+} & \bf {20}\\
\hline
\bf {A C P}  & \bf {28.04} & \bf {5.59} & \bf {21.15} & \bf {18.26} & \bf {+} & \bf {29.67} & \bf {4.11} & \bf {20.23} & \bf {18} & \bf {+} & \bf {21}\\
\hline
A L P  & 26.27 & 4.69 & 19.55 & 16.84 & - & 30.04 & 3.59 & 23.13 & 18.92 & + & 22\\
\hline
A L C & 26.78 & 7.46 & 20.62 & 18.29 & + & 24.84 & 3.84 & 21.33 & 16.67 & - & 23\\
\hline
\bf ALL & \bf 25.71 & \bf 6.39 & \bf 21.31 & \bf 17.8 & \bf + & \bf 28.31 & \bf 5.3 & \bf 21.89 & \bf 18.5 & \bf + & \bf 24\\
\hline

\end{tabular}
\end{center}
\caption{\label{tab:FullBERTmodels_results} Full ROUGE results of BERT multitask model. $\Delta$ represents the change direction relative to the abstractive only model, where 
'+' means higher average ROUGE, and '-' otherwise. \textbf{Boldface} indicates improving scores across all courses.}
\end{table*}


\begin{table*}[ht]
\small
\begin{tabular}{|l|c|c|c|c|c||c|c|c|c|c|c|}
\hline
\multirow{2}{*}{Tasks}               & R1       & R2       & RL    & AVG   & $\Delta$ & R1        & R2       & RL & AVG & $\Delta$  &   \\
\cline{2-12}
& \multicolumn{5}{c|}{\textbf{CS0445}}   & \multicolumn{5}{c|}{\textbf{ENGR}} & \\
\hline
Single Task (Abs.)  & 34.62   & 9.46    & 29.84  &  24.64  & * & 35.43    & 9.93   & 31.07 & 25.47 & * &1\\
\hline
A E     &   30.01             & 8.21         &    22.92      &  20.38     & - &    32.04    & 8.11    & 27.10  & 22.41 & - & 2\\
\hline 
A C &    34.42  & 9.71 & 29.31 & 24.48  & - & 35.84 &  10.14  &31.38  & 25.78 &+&  3\\
\hline
{\textbf{{A P}}}              &   {{34.56}} &  {\textbf{{9.81}}}   & {\textbf{{30.11}}}    & {\textbf{{24.82}}}  & {\textbf{{+}}} & {\textbf{{36.79}}} &  {\textbf{{12.64}}}     & {\textbf{{32.62}}} & {\textbf{{27.35}}} & {\textbf{{+}}} &  {\textbf{{4}}}\\
\hline
A C P   &    34.70      &   9.47       & 30.2        & 27.79 &  +  &    36.16    &  11.46        &  31.74       & 26.45 &+ &  5\\
\hline
A E C & 27.43&  7.54& 24.63& 19.86 &- &  29.41 & 7.63  & 26.15 & 21.06  & -  & 6\\
\hline
ALL               &28.34  &   8.31 &26.72  & 21.12 & -   &    30.11    & 8.45& 28.98 & 22.51& - & 7\\
\hline
\hline
& \multicolumn{5}{c|}{\textbf{S2015}} & \multicolumn{5}{c|}{\textbf{S2016}} & \\ \hline
Single Task (Abs.)    & 36.87   & 12.03   & 32.34   & 27.08 & * & 37.41  &12.33   &33.02   & 27.58 & * &  12\\
\hline
A E  &27.65         & 7.96       &22.74    & 19.45  & - & 30.25          &   10.93       &  26.45   & 22.54 & - & 13 \\ 
\hline 
A C & 34.49        & 10.40 & 30.12        & 25 &   -    &  37.09     &   12.77       &    32.42     & 27.42 & - & 14\\
\hline
{\textbf{{A P}}}       &    {{36.78}} &  {\textbf{{12.64}}}  & {\textbf{{32.62}}}    & {\textbf{{27.34}}} & {\textbf{{+}}}  & {\textbf{{38.86}}}  & {\textbf{{13.41}}}  &  {\textbf{{33.84}}}  & {\textbf{{28.71}}} & {\textbf{{+}}} & {\textbf{{15}}} \\
\hline
A C P &     35.63  &  11.14       &  30.85 &25.87 & - &   38.64     &       14.27 &   33.52  &28.81 & + & 16\\
\hline
A E C& 28.25 &7.97 &23.15 & 19.79 & - &  31.65 &11.60  &26.86  & 23.37  & - & 17\\
\hline
ALL     & 31.21   & 10.66 &28.99  & 23.62 &-  & 31.57           & 10.99         &27.20         &23.25 & - & 18\\
\hline
\end{tabular}

\caption{\label{tab:FullT5models_results} Full ROUGE results of T5 Model fine-tuned on CM data under several experimentation settings}
\end{table*}

\end{document}